\documentclass[11pt,a4paper]{article}
\usepackage[hyperref]{acl2017}
\usepackage{latexsym}

\usepackage{url}

\usepackage{amsmath}
\usepackage{tikz}
\usepackage{subfigure}
\usepackage{multirow}

\usepackage{txfonts}
\aclfinalcopy 

\newcommand{\GRU}{\ensuremath{\mathrm{GRU}}}
\newcommand*{\affaddr}[1]{#1}
\newcommand*{\affmark}[1][*]{\textsuperscript{#1}}

\newcommand*{\email}[1]{\texttt{#1}}

\title{Improved Neural Machine Translation with a \\ Syntax-Aware Encoder and Decoder}

\author{%
Huadong Chen\affmark[\dag], Shujian Huang\affmark[\dag]\thanks{\quad Corresponding author.}, David Chiang\affmark[\ddag], Jiajun Chen\affmark[\dag]\\
\affaddr{\affmark[\dag]State Key Laboratory for Novel Software Technology, Nanjing University}\\
\email{\{chenhd,huangsj,chenjj\}@nlp.nju.edu.cn}\\
\affaddr{\affmark[\ddag]Department of Computer Science and Engineering, University of Notre Dame}\\
\email{ dchiang@nd.edu}\\
}

\date{}

\begin{document}
\maketitle
\begin{abstract}
Most neural machine translation (NMT) models are based on the sequential encoder-decoder framework, which makes no use of syntactic information. In this paper, we improve this model by explicitly incorporating source-side syntactic trees. More specifically, we propose (1) a \emph{bidirectional tree encoder} which learns both sequential and tree structured representations; (2) a \emph{tree-coverage model} that lets the attention depend on the source-side syntax.
Experiments on Chinese-English translation demonstrate that our proposed models outperform the sequential attentional model as well as a stronger baseline with a bottom-up tree encoder and word coverage.\footnote{Our code is publicly available at \url{https://github.com/howardchenhd/Syntax-awared-NMT/}}
\end{abstract}

\section{Introduction}
Recently, neural machine translation (NMT) models \cite{NIPS2014_5346,bahdanau2014neural} have obtained state-of-the-art performance on many language pairs. Their success depends on the representation they use to bridge the source and target language sentences. However, this representation, a sequence of fixed-dimensional vectors, differs considerably from most theories about mental representations of sentences, and from traditional natural language processing pipelines, in which semantics is built up compositionally using a recursive syntactic structure.

Perhaps as evidence of this, current NMT models still suffer from syntactic errors such as attachment~\cite{shi-padhi-knight:2016:EMNLP2016}. We argue that instead of letting the NMT model rely solely on the implicit structure it learns during training~\cite{DBLP:journals/corr/ChoMBB14}, we can improve its performance by augmenting it with explicit structural information and using this information throughout the model. This has two benefits.

\begin{figure}
    \centering
    \subfigure[example sentence pair with alignments]{
    \includegraphics[width=0.95\hsize]{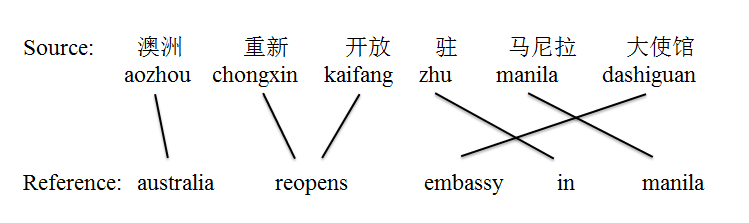}
    \label{fig:sub_pair}
    }
    \quad

    \subfigure[binarized source side tree]{
    \begin{tikzpicture}
    \tikzset{every node/.style={align=center,anchor=north}}
    \node(u) at (-1,0) {{\tiny aozhou} \\$x_1$};
    \node(a) at (0,0) {{\tiny chongxin}\\$x_2$};
    \node(b) at (1,0) {{\tiny kaifang}\\$x_3$};
    \node(c) at (2,0){{\tiny zhu}\\$x_4$};
    \node(d) at (3,0){{\tiny manila}\\$x_5$};
    \node(e) at (4,0){{\tiny dashiguan}\\$x_6$};

    \draw[-](c.north)--(2.5,0.5);
    \draw[-](d.north)--(2.5,0.5);
    \draw[-](2.5,0.5)--(3,1);
    \draw[-](e.north)--(3,1);
    \draw[-](3,1)--(2,1.5);
    \draw[-](b.north)--(2,1.5);
    \draw[-](2,1.5)--(1,2);
    \draw[-](a.north)--(1,2);

    \draw[-](1,2)--(0,2.5);
    \draw[-](u.north)--(0,2.5);
    \end{tikzpicture}
    \label{fig:sub1}
    } 
    \label{fg_sent}
    \caption{An example sentence pair (a), with its binarized source side tree (b). We use $x_{i}$ to represent the $i$-th word in the source sentence. We will use this sentence pairs as the running example throughout this paper.}
\end{figure}
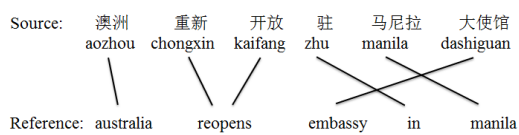
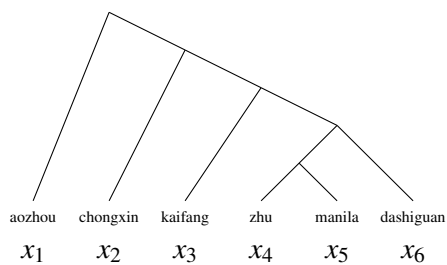

First, the explicit syntactic information will help the encoder generate better source side representations. \newcite{li-EtAl:2015:EMNLP5} show that for tasks in which long-distance semantic dependencies matter, representations learned from recursive models using syntactic structures may be more powerful than those from sequential recurrent models. In the NMT case, given syntactic information, it will be easier for the encoder to incorporate long distance dependencies into better representations, which is especially important for the translation of long sentences.

Second, it becomes possible for the decoder to use syntactic information to guide its reordering decisions better (especially for language pairs with significant reordering, like Chinese-English). Although the attention model~\cite{bahdanau2014neural} and the coverage model~\cite{tu-EtAl:2016:P16-1,mi-EtAl:2016:EMNLP2016} provide effective mechanisms to control the generation of translation, these mechanisms work at the word level and cannot capture phrasal cohesion between the two languages~\cite{Fox:2002:PCS:1118693.1118732,yoonkim}. With explicit syntactic structure, the decoder can generate the translation more in line with the source syntactic structure. For example, when translating the phrase \emph{zhu manila dashiguan} in Figure~\ref{fg_sent}, the tree structure indicates that \emph{zhu}~`in' and \emph{manila} form a syntactic unit, so that the model can avoid breaking this unit up to make an incorrect translation like ``in embassy of manila''~\footnote{According to the source sentence, ``embassy'' belongs to ``australia'', not ``manila''.}.

In this paper, we propose a novel encoder-decoder model that makes use of a precomputed source-side syntactic tree in both the encoder and decoder. In the encoder (\S\ref{sect:bi-tree}), we improve the tree encoder of \newcite{eriguchi-hashimoto-tsuruoka:2016:P16-1} by introducing a \emph{bidirectional tree encoder}. For each source tree node (including the source words), we generate a representation containing  information both from below (as with the original bottom-up encoder) and from above (using a top-down encoder). Thus, the annotation of each node summarizes the surrounding sequential context, as well as the entire syntactic context.

In the decoder (\S\ref{sect:structural}), we incorporate source syntactic tree structure into the attention model via an extension of the coverage model of \citet{tu-EtAl:2016:P16-1}. With this \emph{tree-coverage model}, we can better guide the generation phase of translation, for example, to learn a preference for phrasal cohesion~\cite{Fox:2002:PCS:1118693.1118732}. Moreover, with a tree encoder, the decoder may try to translate both a parent and a child node, even though they overlap; the tree-coverage model enables the decoder to learn to avoid this problem.

To demonstrate the effectiveness of the proposed model, we carry out experiments on Chinese-English translation. Our experiments show that: (1) our bidirectional tree encoder based NMT system achieves significant improvements over the standard attention-based NMT system, and (2) incorporating source tree structure into the attention model yields a further improvement. In all, we demonstrate an improvement of~+3.54 BLEU over a standard attentional NMT system, and~+1.90 BLEU over a stronger NMT system with a Tree-LSTM encoder~\cite{eriguchi-hashimoto-tsuruoka:2016:P16-1} and a coverage model~\cite{tu-EtAl:2016:P16-1}. To the best of our knowledge, this is the first work that uses source-side syntax in both the encoder and decoder of an NMT system.

\section{Neural Machine Translation}

Most NMT systems follow the encoder-decoder framework with attention, first proposed by \newcite{bahdanau2014neural}. Given a source sentence $\mathbf{x}=x_1 \cdots x_i \cdots x_I$ and a target sentence $\mathbf{y}=y_1 \cdots y_j \cdots y_J$, NMT aims to directly model the translation probability:
\begin{equation}
	P(\mathbf{y}\mid\mathbf{x}; \theta)= \prod_{1}^{J}{P(y_{j}\mid\mathbf{y_{<j}}, \mathbf{x}; \theta )},
\end{equation}
where $\theta$ is a set of parameters and $y_{<j}$ is the sequence of previously generated target words.
Here, we briefly describe the underlying framework of the encoder-decoder NMT system.

 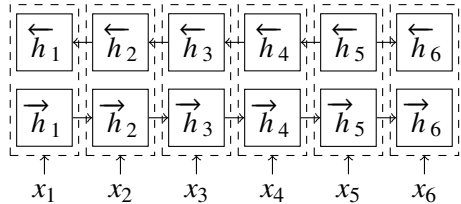
\begin{figure}
     \centering
     \begin{tikzpicture}

    \node[](a) at (0,0) {$x_1$};
    \node[](b) at (1,0) {$x_2$};
    \node[](c) at (2,0){$x_3$};
    \node[](d) at (3,0){$x_4$};
    \node[](e) at (4,0){$x_5$};
    \node[](f) at (5,0) {$x_6$};

    \node[draw,rectangle](a1) at (0,1) {$\overrightarrow{h}_1$};
    \node[draw,rectangle](b1) at (1,1) {$\overrightarrow{h}_2$};
    \node[draw,rectangle](c1) at (2,1){$\overrightarrow{h}_3$};
    \node[draw,rectangle](d1) at (3,1){$\overrightarrow{h}_4$};
    \node[draw,rectangle](e1) at (4,1){$\overrightarrow{h}_5$};
    \node[draw,rectangle](f1) at (5,1){$\overrightarrow{h}_6$};

    \node[draw,rectangle](a2) at (0,2) {$\overleftarrow{h}_1$};
    \node[draw,rectangle](b2) at (1,2) {$\overleftarrow{h}_2$};
    \node[draw,rectangle](c2) at (2,2){$\overleftarrow{h}_3$};
    \node[draw,rectangle](d2) at (3,2){$\overleftarrow{h}_4$};
    \node[draw,rectangle](e2) at (4,2){$\overleftarrow{h}_5$};
    \node[draw,rectangle](f2) at (5,2){$\overleftarrow{h}_6$};

    \draw [dashed]  (-0.45,0.5) rectangle (0.45,2.5);
    \draw [dashed]  (0.55,0.5) rectangle (1.45,2.5);
    \draw [dashed]  (1.55,0.5) rectangle (2.45,2.5);
    \draw [dashed]  (2.55,0.5) rectangle (3.45,2.5);
    \draw [dashed]  (3.55,0.5) rectangle (4.45,2.5);
    \draw [dashed]  (4.55,0.5) rectangle (5.45,2.5);

    \draw[->](a1)--(b1);
    \draw[->](b1)--(c1);
    \draw[->](c1)--(d1);
    \draw[->](d1)--(e1);
    \draw[->](e1)--(f1);

    \draw[->](e2)--(d2);
    \draw[->](d2)--(c2);
    \draw[->](c2)--(b2);
    \draw[->](b2)--(a2);
    \draw[->](e2)--(f2);

    \draw[->](a)--(0,0.5);
    \draw[->](b)--(1,0.5);
    \draw[->](c)--(2,0.5);
    \draw[->](d)--(3,0.5);
    \draw[->](e)--(4,0.5);
    \draw[->](f)--(5,0.5);

    \end{tikzpicture}
    \caption{Illustration of the bidirectional sequential encoder. The dashed rectangle represents the annotation of word $x_i$.}
    \label{fig:encoder}
 \end{figure}

\subsection{Encoder Model}
Following~\citet{bahdanau2014neural}, we use a bidirectional gated recurrent unit (GRU)~\cite{cho2014learning} to encode the source sentence, so that the annotation of each word contains a summary of both the preceding and following words. The bidirectional GRU consists of a forward and a backward GRU, as shown in Figure~\ref{fig:encoder}. The forward GRU reads the source sentence from left to right and calculates a sequence of forward hidden states $(\overrightarrow{h_1},\dots,\overrightarrow{h_I})$. The backward GRU scans the source sentence from right to left, resulting in a
sequence of backward hidden states $(\overleftarrow{h_1},\dots,\overleftarrow{h_I})$. Thus
\begin{equation}
   \begin{split}
	\overrightarrow{h_i} &= \GRU(\overrightarrow{h_{i-1}}, s_i) \\
	\overleftarrow{h_i} &= \GRU(\overleftarrow{h_{i-1}}, s_i)
	\end{split}
\end{equation}
where $s_i$ is the $i$-th source word's word embedding, and $\text{GRU}$ is a gated recurrent unit; see the paper by \citet{cho2014learning} for a definition.

The \emph{annotation} of each source word $x_i$ is obtained by concatenating the forward and backward hidden states:
\[\overleftrightarrow{h_i} = \begin{bmatrix} \overrightarrow{h_i} \\ \overleftarrow{h_i} \end{bmatrix}.\]  The whole sequence of these annotations is used by the decoder.

\subsection{Decoder Model}
The decoder is a forward GRU predicting the translation $\mathbf{y}$ word by word. The probability of generating the $j$-th word $y_j$ is:
\begin{equation}
P(y_{j}\mid\mathbf{y_{<j}}, \mathbf{x}; \theta) = \text{softmax}(t_{j-1}, d_{j}, c_{j})
\end{equation}
where
$t_{j-1}$ is the word embedding of the $(j-1)$-th target word, $d_{j}$ is the decoder's hidden state of time $j$, and $c_{j}$ is the \emph{context vector} at time $j$. The state $d_{j}$ is computed as
\begin{equation}
	d_j = \GRU(d_{j-1}, t_{j-1}, c_j),
\end{equation}
where $\GRU(\mathord\cdot)$ is extended to more than two arguments by first concatenating all arguments except the first.

The attention mechanism computes the context vector $c_i$ as a weighted sum of the source annotations,
\begin{equation}
    c_{j} = \sum_{i=1}^{I}{\alpha_{j,i}\overleftrightarrow{h_{i}}}
\end{equation}
where the attention weight $\alpha_{j,i}$ is
\begin{align}
\alpha_{j,i} = \frac{\exp{(e_{j,i})}}{\sum_{i'=1}^{I}\exp{(e_{j,i'})}}
\end{align}
and
\begin{align}\label{eq_att}
 e_{j,i} = v_{a}^\mathrm{T} \tanh{(W_ad_{j-1} + U_a\overleftrightarrow{h_i})}
\end{align}
where $v_a$, $W_a$ and $U_a$ are the weight matrices of the attention model, and $e_{j,i}$ is an attention model that scores how well $d_{j-1}$ and $\overleftrightarrow{h_i}$ match.

With this strategy, the decoder can attend to the source annotations that are most relevant at a given time.

\section{Tree Structure Enhanced Neural Machine Translation}\label{sec:length}
Although syntax has shown its effectiveness in non-neural statistical machine translation (SMT) systems~\cite{Yamada2001,Koehn:2003:SPT:1073445.1073462,Liu2006,Chiang:2007:HPT:1268656.1268659}, most proposed NMT models (a notable exception being that of~\newcite{eriguchi-hashimoto-tsuruoka:2016:P16-1})
process a sentence only as a sequence of words, and do not explicitly exploit the inherent structure of natural language sentences. In this section, we present models which directly incorporate source syntactic trees into the encoder-decoder framework.

\subsection{Preliminaries}

Like~\newcite{eriguchi-hashimoto-tsuruoka:2016:P16-1}, we currently focus on source side syntactic trees, which can be computed prior to translation. Whereas~\newcite{eriguchi-hashimoto-tsuruoka:2016:P16-1} use HPSG trees, we use phrase-structure trees as in the Penn Chinese Treebank~\cite{Xue:2005:PCT:1064781.1064785}. Currently, we are only using the structure information from the tree without the syntactic labels. Thus our approach should be applicable to any syntactic grammar that provides such a tree structure (Figure~\ref{fig:sub1}).

More formally, the encoder is given a source sentence $\mathbf{x} = x_1 \cdots x_I$ as well as a source tree whose leaves are labeled $x_1, \ldots, x_I$. We assume that this tree is strictly binary branching. For convenience, each node is assigned an index. The leaf nodes get indices $1, \ldots, I$, which is the same as their word indices. For any node with index $k$, let $p(k)$ denote the index of the node's parent (if it exists), and $L(k)$ and $R(k)$ denote the indices of the node's left and right children (if they exist).

\subsection{Tree-GRU Encoder}\label{sect:uni-tree}

We first describe tree encoders~\citep{tai-socher-manning:2015:ACL-IJCNLP,eriguchi-hashimoto-tsuruoka:2016:P16-1}, and then discuss our improvements.

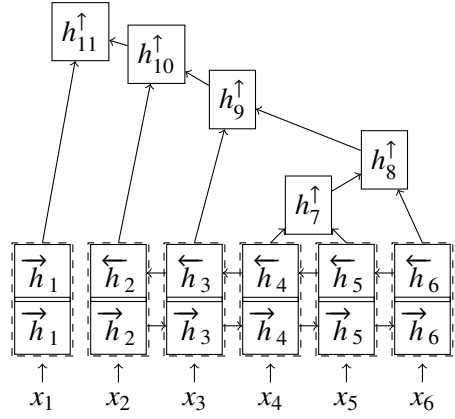
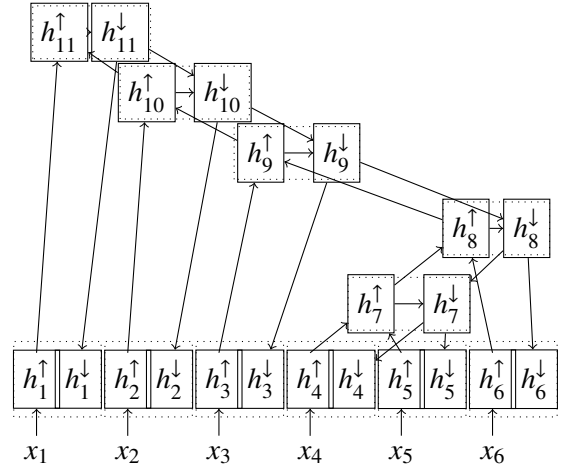
\begin{figure}[ht]
    \centering

    \subfigure[Tree-GRU Encoder]{
    \begin{tikzpicture}
    \node[](u) at (-1,0) {$x_1$};
    \node[](a) at (0,0) {$x_2$};
    \node[](b) at (1,0) {$x_3$};
    \node[](c) at (2,0){$x_4$};
    \node[](d) at (3,0){$x_5$};
    \node[](e) at (4,0){$x_6$};

    \node[draw,rectangle](u1) at (-1,1) {$\overrightarrow{h}_1$};
    \node[draw,rectangle](a1) at (0,1) {$\overrightarrow{h}_2$};
    \node[draw,rectangle](b1) at (1,1) {$\overrightarrow{h}_3$};
    \node[draw,rectangle](c1) at (2,1){$\overrightarrow{h}_4$};
    \node[draw,rectangle](d1) at (3,1){$\overrightarrow{h}_5$};
    \node[draw,rectangle](e1) at (4,1){$\overrightarrow{h}_6$};

    \node[draw,rectangle](u1) at (-1,1.7) {$\overrightarrow{h}_1$};
    \node[draw,rectangle](a2) at (0,1.7) {$\overleftarrow{h}_2$};
    \node[draw,rectangle](b2) at (1,1.7) {$\overleftarrow{h}_3$};
    \node[draw,rectangle](c2) at (2,1.7){$\overleftarrow{h}_4$};
    \node[draw,rectangle](d2) at (3,1.7){$\overleftarrow{h}_5$};
    \node[draw,rectangle](e2) at (4,1.7){$\overleftarrow{h}_6$};

    \draw [dashed]  (-1.4,0.6) rectangle (-0.6,2.1);
    \draw [dashed]  (-0.4,0.6) rectangle (0.4,2.1);
    \draw [dashed]  (0.6,0.6) rectangle (1.4,2.1);
    \draw [dashed]  (1.6,0.6) rectangle (2.4,2.1);
    \draw [dashed]  (2.6,0.6) rectangle (3.4,2.1);
    \draw [dashed]  (3.6,0.6) rectangle (4.4,2.1);

    \draw[->](a1)--(b1);
    \draw[->](b1)--(c1);
    \draw[->](c1)--(d1);
    \draw[->](d1)--(e1);

    \draw[->](e2)--(d2);
    \draw[->](d2)--(c2);
    \draw[->](c2)--(b2);
    \draw[->](b2)--(a2);

    \draw[->](u)--(-1,0.5);
    \draw[->](a)--(0,0.5);
    \draw[->](b)--(1,0.5);
    \draw[->](c)--(2,0.5);
    \draw[->](d)--(3,0.5);
    \draw[->](e)--(4,0.5);

    \node[draw,rectangle](u1) at (2.5, 2.6) {$h^{\uparrow}_7$};

    \node[draw,rectangle](u2) at (3.5, 3.2) {$h^{\uparrow}_8$};

    \draw[->](2,2.1)--(u1);
    \draw[->](3,2.1)--(u1);
    \draw[->](u1)--(u2);
    \draw[->](4,2.1)--(u2);

    \node[draw,rectangle](u3) at (1.5, 4) {$h^{\uparrow}_9$};

    \draw[->](u2)--(u3);
    \draw[->](1,2.1)--(u3);

    \node[draw,rectangle](u4) at (0.5, 4.6) {$h^{\uparrow}_{10}$};

    \draw[->](u3)--(u4);
    \draw[->](0,2.1)--(u4);

     \node[draw,rectangle](u5) at (-0.5, 4.9) {$h^{\uparrow}_{11}$};

    \draw[->](u4)--(u5);
    \draw[->](-1,2.1)--(u5);

    \end{tikzpicture}
    \label{fig:sub3}
    }

    \quad

    \subfigure[Bidirectional Tree Encoder]{
    \begin{tikzpicture}

    \node[](x) at (-1.2,0) {$x_1$};

    \node[](a) at (0,0) {$x_2$};
    \node[](b) at (1.2,0) {$x_3$};
    \node[](c) at (2.4,0){$x_4$};
    \node[](d) at (3.6,0){$x_5$};
    \node[](e) at (4.8,0){$x_6$};

    \node[draw,rectangle](x1) at (-1.2,1) {$h^{\uparrow}_1$};

    \node[draw,rectangle](a1) at (0,1) {$h^{\uparrow}_2$};
    \node[draw,rectangle](b1) at (1.2,1) {$h^{\uparrow}_3$};
    \node[draw,rectangle](c1) at (2.4,1){$h^{\uparrow}_4$};
    \node[draw,rectangle](d1) at (3.6,1){$h^{\uparrow}_5$};
    \node[draw,rectangle](e1) at (4.8,1){$h^{\uparrow}_6$};

    \node[draw,rectangle](x2) at (-0.65,1) {$h^{\downarrow}_1$};

    \node[draw,rectangle](a2) at (0.55,1) {$h^{\downarrow}_2$};
    \node[draw,rectangle](b2) at (1.75,1) {$h^{\downarrow}_3$};
    \node[draw,rectangle](c2) at (2.95,1){$h^{\downarrow}_4$};
    \node[draw,rectangle](d2) at (4.15,1){$h^{\downarrow}_5$};
    \node[draw,rectangle](e2) at (5.35,1){$h^{\downarrow}_6$};

    \draw [dotted]  (-1.5,0.5) rectangle (-0.35,1.5);
    \draw [dotted]  (-0.3,0.5) rectangle (0.85,1.5);
    \draw [dotted]  (0.9,0.5) rectangle (2.05,1.5);
    \draw [dotted]  (2.1,0.5) rectangle (3.25,1.5);
    \draw [dotted]  (3.3,0.5) rectangle (4.45,1.5);
    \draw [dotted]  (4.5,0.5) rectangle (5.65,1.5);

    \draw[->](x)--(x1);
    \draw[->](a)--(a1);
    \draw[->](b)--(b1);
    \draw[->](c)--(c1);
    \draw[->](d)--(d1);
    \draw[->](e)--(e1);

    \node[draw,rectangle](u1) at (3.2, 2) {$h^{\uparrow}_7$};
    \node[draw,rectangle](d6) at (4.2, 2) {$h^{\downarrow}_7$};

    \draw[->](u1)--(d6);

    \draw [dotted]  (2.9,1.65) rectangle (4.5,2.35);

    \draw[->](2.4,1.35)--(u1);
    \draw[->](3.6,1.35)--(u1);

    \node[draw,rectangle](u2) at (4.45, 3) {$h^{\uparrow}_8$};
    \node[draw,rectangle](d7) at (5.25, 3) {$h^{\downarrow}_8$};

    \draw[->](u2)--(d7);

    \draw [dotted]  (4.15,2.65) rectangle (5.55,3.35);

    \draw[->](u1)--(u2);
    \draw[->](4.8,1.35)--(u2);

    \node[draw,rectangle](u3) at (1.75, 4) {$h^{\uparrow}_9$};
    \node[draw,rectangle](d8) at (2.75, 4) {$h^{\downarrow}_9$};

    \draw[->](u3)--(d8);

    \draw [dotted]  (1.4,3.65) rectangle (3.05,4.35);

    \draw[->](u2)--(u3);
    \draw[->](1.2,1.35)--(u3);

    \node[draw,rectangle](u4) at (0.25, 4.8) {$h^{\uparrow}_{10}$};
    \node[draw,rectangle](d9) at (1.25, 4.8) {$h^{\downarrow}_{10}$};

    \draw[->](u4)--(d9);

    \draw [dotted]  (-0.1,4.45) rectangle (1.6,5.15);

    \node[draw,rectangle](u5) at (-0.9, 5.6) {$h^{\uparrow}_{11}$};
    \node[draw,rectangle](d10) at (-0.1, 5.6) {$h^{\downarrow}_{11}$};

    \draw[->](u5)--(d10);

    \draw [dotted]  (-1.25,5.25) rectangle (0.3,5.95);

    \draw[->](u4)--(u5);
    \draw[->](-1.2,1.35)--(u5);
    \draw[->](d10)--(x2);

    \draw[->](u3)--(u4);
    \draw[->](0,1.35)--(u4);

    \draw[->](d10)--(d9);
    \draw[->](d9)--(d8);
    \draw[->](d8)--(d7);
    \draw[->](d7)--(d6);

    \draw[->](d9)--(a2);
    \draw[->](d8)--(b2);
    \draw[->](d7)--(e2);
    \draw[->](d6)--(c2);
    \draw[->](d6)--(d2);

    \end{tikzpicture}
    \label{fig:sub4}
    }

    \quad

    \caption{Illustration of the proposed encoder models for the running example. The non-leaf nodes are assigned with index 7-11. The annotations $h^{\uparrow}_i$ of leaf nodes in (b) are identical to the annotations (dashed rectangles) of leaf nodes in (a). The dotted rectangles in (b) indicate the annotation produced by the bidirectional tree encoder.}
    \label{fig:my_label}
\end{figure}

Following~\newcite{eriguchi-hashimoto-tsuruoka:2016:P16-1}, we build a tree encoder on top of the sequential encoder~(as shown in Figure~\ref{fig:sub3}). If node $k$ is a leaf node, its hidden state is the annotation produced by the sequential encoder:
\begin{equation*}
h^\uparrow_k = \overleftrightarrow{h_k}.
\end{equation*}
Thus, the encoder is able to capture both sequential context and syntactic context.

If node $k$ is an interior node, its hidden state is the combination of its previously calculated left child hidden state $h_{L(k)}$ and right child hidden state $h_{R(k)}$:
\begin{equation}\label{eq_t}
h^{\uparrow}_{k} = f(h^\uparrow_{L(k)}, h^\uparrow_{R(k)})
\end{equation}
where $f(\cdot)$ is a nonlinear function, originally a Tree-LSTM~\cite{tai-socher-manning:2015:ACL-IJCNLP,eriguchi-hashimoto-tsuruoka:2016:P16-1}.

The first improvement we make to the above tree encoder is that, to be consistent with the sequential encoder model, we use Tree-GRU units instead of Tree-LSTM units. Similar to Tree-LSTMs, the Tree-GRU has gating mechanisms to control the information flow inside the unit for every node without separate memory cells. Then, Eq.~\ref{eq_t} is calculated by a Tree-GRU as follows:
\begin{align*}
     r_L &= \sigma{(U^{(rL)}_Lh^\uparrow_{L(k)} + U^{(rL)}_Rh^\uparrow_{R(k)} + b^{(rL)})}  \\
     r_R &= \sigma(U^{(rR)}_Lh^\uparrow_{L(k)} + U^{(rR)}_Rh^\uparrow_{R(k)} + b^{(rR)})   \\
     z_L &= \sigma(U^{(zL)}_Lh^\uparrow_{L(k)} + U^{(zL)}_Rh^\uparrow_{R(k)} + b^{(zL)})   \\
     z_R &= \sigma(U^{(zR)}_Lh^\uparrow_{L(k)} + U^{(zR)}_Rh^\uparrow_{R(k)} + b^{(zR)})   \\
     z &= \sigma(U^{(z)}_Lh^\uparrow_{L(k)} + U^{(z)}_Rh^\uparrow_{R(k)} + b^{(z)})  \\
     \tilde{h}^{\uparrow}_{k} &= \tanh{\left(U_L(r_L \odot h^\uparrow_{L(k)}) + U_R(r_R \odot h^\uparrow_{R(k)})\right)} \\
     h^{\uparrow}_{k} &= z_L \odot h^\uparrow_{L(k)} + z_R \odot h^\uparrow_{R(k)} + z \odot \tilde{h}^{\uparrow}_{k}
\end{align*}
where $r_L, r_R$ are the reset gates and $z_L, z_R$ are the update gates for the left and right children, and $z$ is the update gate for the internal hidden state $\tilde{h}^{{\uparrow}}_{k}$. The $U^{(\cdot)}$ and~$b^{(\cdot)}$ are the weight matrices and bias vectors.

\subsection{Bidirectional Tree Encoder}\label{sect:bi-tree}

Although the bottom-up tree encoder can take advantage of syntactic structure, the learned representation of a node is based on its subtree only; it contains no information from higher up in the tree. In particular, the representation of leaf nodes is still the sequential one. Thus no syntactic information is fed into words. By analogy with the bidirectional sequential encoder, we propose a natural extension of the bottom-up tree encoder: the bidirectional tree encoder (Figure~\ref{fig:sub4}).

Unlike the bottom-up tree encoder or the right-to-left sequential encoder, the top-down encoder by itself would have no lexical information as input. To address this issue, we feed the hidden states of the bottom-up encoder to the top-down encoder. In this way, the information of the whole syntactic tree is handed to the root node and propagated to its offspring by the top-down encoder.

In the top-down encoder, each hidden state has only one predecessor. In fact, the top-down path from root of a tree to any node can be viewed as a sequential recurrent neural network. We can calculate the hidden states of each node top-down using a standard sequential GRU.

First, the hidden state of the root node $\rho$ is simply computed as follows:
\begin{equation}
h^{\downarrow}_{\rho} = \tanh{(Wh^{\uparrow}_{\rho} + b)}
\end{equation}
where $W$ and $b$ are a weight matrix and bias vector.

Then, other nodes are calculated by a $GRU$. For hidden state $h^{\downarrow}_{k}$:
\begin{equation}
h^{\downarrow}_{k} = \GRU(h^{\downarrow}_{p(k)}, h^{\uparrow}_{k})
\end{equation}
where $p(k)$ is the parent index of $k$. We replace the weight matrices $W^{r}$, $U^{r}$, $W^{z}$, $U^{z}$, $W$ and $U$ in the standard $\GRU$ with $P^{r}_{D}$, $Q^{r}_{D}$, $P^{z}_{D}$, $Q^{z}_{D}$, $P_{D}$, and $Q_{D}$, respectively. The subscript $D$ is either $L$ or $R$ depending on whether node $k$ is a left or right child, respectively.

Finally, the annotation of each node is obtained by concatenating its bottom-up hidden state and top-down hidden state:
\begin{equation*}
h^\updownarrow_k = \begin{bmatrix} h^{\uparrow}_k \\[1ex]  h^{\downarrow}_k\end{bmatrix}.
\end{equation*}
This allows the tree structure information flow from the root to the leaves (words). Thus, all the annotations are based on the full context of word sequence and syntactic tree structure.

\newcite{2017arXiv170101811K} propose a similar bidirectional Tree-GRU for sentiment analysis, which differs from ours in several respects: in the bottom-up encoder, we use separate reset/update gates for left and right children, analogous to Tree-LSTMs \cite{tai-socher-manning:2015:ACL-IJCNLP}; in the top-down encoder, we use separate weights for left and right children.

\newcite{DBLP:journals/corr/TengZ16} also propose a  bidirectional Tree-LSTM encoder for classification tasks. They use a more complex head-lexicalization scheme to feed the top-down encoder. We will compare their model with ours in the experiments.

\subsection{Tree-Coverage Model}\label{sect:structural}

\begin{figure}[!t]\footnotesize
	\centering
	
	\subfigure[Tree-GRU Encoder]{
			\includegraphics[width=3in]{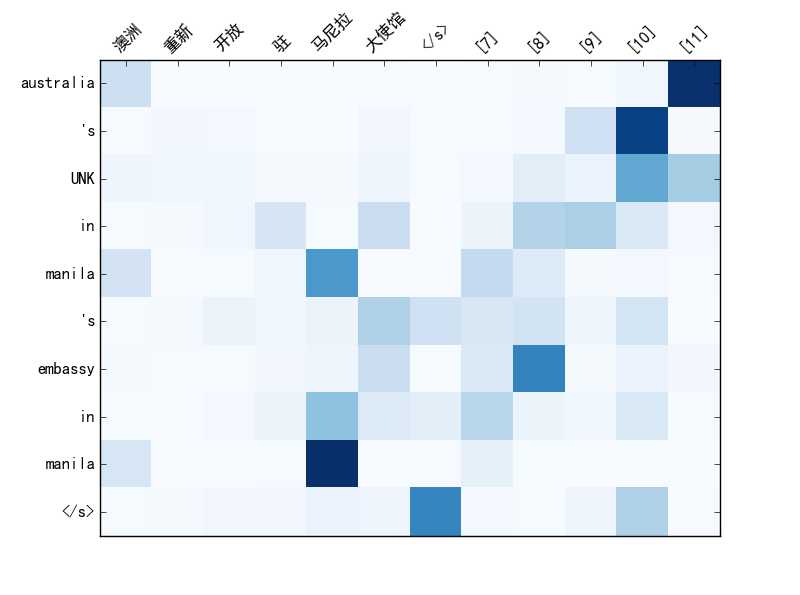}
			\label{sub_att}
			}
	\quad
	\subfigure[+ Tree-Coverage Model]{
			\includegraphics[width=3in]{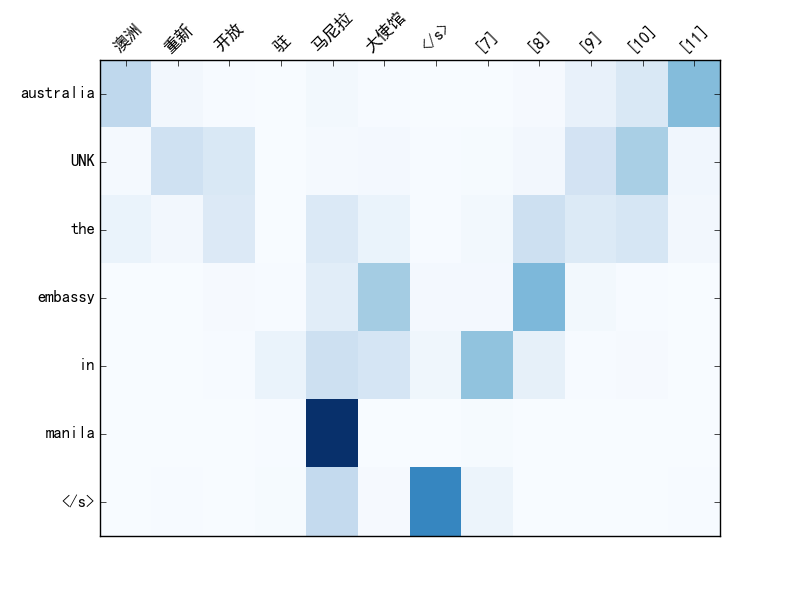}
			\label{sub_att2}
			}

	\caption{The attention heapmap plotting the attention weights during different translation steps, for translating the sentence in Figure~\ref{fig:sub_pair}. The nodes $[7]$-$[11]$ correspond to non-leaf nodes indexed in Figure~\ref{fig:my_label}. Incorporating Tree-Coverage Model produces more concentrated alignments and alleviates the over-translation problem.}
	
	\label{fig_attention}
	
\end{figure}

We also extend the decoder to incorporate information about the source syntax into the attention model. We have observed two issues in translations produced using the tree encoder. First, a syntactic phrase in the source sentence is often incorrectly translated into discontinuous words in the output. Second, since the non-leaf node annotations contain more information than the leaf node annotations, the attention model prefers to attend to the non-leaf nodes, which may aggravate the over-translation problem (translating the same part of the sentence more than once).

As shown in Figure~\ref{sub_att}, almost all the non-leaf nodes are attended too many times during decoding. As a result, the Chinese phrase \emph{zhu manila} is translated twice because the model attends to the node spanning \emph{zhu manila} even though both words have already been translated; there is no mechanism to prevent this.

Inspired by the approaches of~\newcite{cohn-EtAl:2016:N16-1}, ~\newcite{feng-EtAl:2016:COLING3}, \newcite{tu-EtAl:2016:P16-1} and \newcite{mi-EtAl:2016:EMNLP2016}, we propose to use prior knowledge to control the attention mechanism. In our case, the prior knowledge is the source syntactic information.

In particular, we build our model on top of the word coverage model proposed by \newcite{tu-EtAl:2016:P16-1},
which alleviate the problems of over-translation and under-translation (failing to translate part of a sentence).
The word coverage model makes the attention at a given time step~$j$ dependent on the attention at previous time steps via
\emph{coverage vectors}:
\begin{equation}
C_{j,i} = \GRU(C_{j-1,i}, \alpha_{j,i}, d_{j-1}, h_{i}).
\end{equation}
The coverage vectors are, in turn, used to update the attention at the next time step, by a small modification to the calculation of $e_{j,i}$ in Eq.~(\ref{eq_att}):
\begin{equation}
e_{j,i} = v_{a}^\mathrm{T} \tanh{(W_a{d}_{j-1} + U_ah_i + V_aC_{j-1,i})}. 
\end{equation}

The word coverage model could be interpreted as a control mechanism for the attention model. Like the standard attention model, this coverage model sees the source-sentence annotations as a bag of vectors; it knows nothing about word order, still less about syntactic structure.

For our model, we extend the word coverage model to coverage on the tree structure by adding a coverage vector for each node in the tree. We further incorporate source tree structure information into the calculation of the coverage vector by requiring each node's coverage vector to depend on its children's coverage vectors and attentions at the previous time step:
\begin{equation}
\begin{split}
   C_{j,i} = \GRU(&C_{j-1,i}, \alpha_{j,i}, d_{j-1}, h_{i},  \\
             &C_{j-1,L(i)}, \alpha_{j,L(i)}, \\ &C_{j-1,R(i)}, \alpha_{j,R(i)}) .
\end{split}
\end{equation}

Although both child and parent nodes of a subtree are helpful for translation, they may supply redundant information. With our mechanism, when the child node is used to produce a translation, the coverage vector of its parent node will reflect this fact, so that the decoder may avoid using the redundant information in the parent node. Figure~\ref{sub_att2} shows a heatmap of the attention of our tree structure enhanced attention model. The attention of non-leaf nodes becomes more concentrated and the over-translation of \emph{zhu manila} is corrected.

\section{Experiments}\label{sect:ex}

\subsection{Data}
We conduct experiments on the NIST Chinese-English translation task. The parallel training data consists of 1.6M sentence pairs extracted from LDC corpora,\footnote{LDC2002E18, LDC2003E14, the Hansards portion of LDC2004T08, and LDC2005T06.} with 46.6M Chinese words and 52.5M English words, respectively. We use NIST MT02 as development data, and  NIST MT03--06 as test data. These data are mostly in the same genre (newswire), avoiding the extra consideration of domain adaptation. Table~\ref{tb-data} shows the statistics of the data sets. The Chinese side of the corpora is word segmented using ICTCLAS.\footnote{\url{http://ictclas.nlpir.org}} We parse the Chinese sentences with the Berkeley Parser\footnote{\url{https://github.com/slavpetrov/berkeleyparser}}~\cite{petrov-klein:2007:main} and binarize the resulting trees following~\newcite{Zhang:2009:TPC:1697236.1697267}. The English side of the corpora is lowercased and tokenized.

\begin{table}[t]
	\centering
	\begin{tabular}{c|c|cc}
		\cline{1-3}
		Data  & Usage & Sents. &  \\ \cline{1-3}
		LDC  & train & 1.6M &  \\ \cline{1-3}	
		MT02   & dev & 878 & \\ \cline{1-3}
		MT03   & test & 919 & \\ \cline{1-3}
		MT04   & test & 1,597 & \\ \cline{1-3}
		MT05   & test & 1,082& \\ \cline{1-3}
		MT06   & test & 1,664& \\ \cline{1-3}
	\end{tabular}
	\caption{Experiment data and statistics.}
	\label{tb-data}
\end{table}

We filter out any translation pairs whose source sentences fail to be parsed.
For efficient training, we also filter out the sentence pairs whose source or target lengths are longer than 50. We use a shortlist of the 30,000 most frequent words in each language to train our models, covering approximately 98.2\% and 99.5\% of the Chinese and English tokens, respectively. All out-of-vocabulary words are mapped to a special symbol \texttt{UNK}.

\begin{table*} 
	\centering
	\begin{tabular}{c|l|c|c|cccc|c}
		\multicolumn{1}{c|} {\textbf{\#}} &{\textbf{Encoder}} &{\textbf{Coverage}} &\textbf{MT02}  &\textbf{MT03} &\textbf{MT04}  &\textbf{MT05}  &\textbf{MT06} &\textbf{Average} \\ \hline
          1 &Sequential    &no       &33.76	&31.88	&33.15	&30.55	&27.47	&30.76     \\
          2 &Tree-LSTM &no       &33.83	&33.15	&33.81	&31.22	&27.86	&31.51(+0.75)      \\
          3 &Tree-GRU  &no       &35.39 &33.62 &35.1 &32.55 &28.26 &32.38(+1.62)  \\
          4 &Bidirectional     &no       &35.52   &33.91	&35.51	&33.34	&29.91	&33.17(+2.41)  \\    \hline
          5 &Sequential    &word        &34.21	&32.73	&34.17	&31.64	&28.29	&31.71(+0.95)    \\
          6 &Tree-LSTM &word      &35.81	&33.62	&34.84	&32.6	&28.52	&32.40(+1.64)   \\
          7 &Tree-GRU  &word      &35.91   &33.71	&35.46	&33.02	&29.14	&32.84(+2.08)    \\
          8 &Bidirectional     &word      &36.14   &35.00	&36.07	&33.74	&30.40	&33.80(+3.04)   \\
          \hline
          9 &Tree-LSTM &tree     &34.97	&33.91	&35.21 &33.08 &	29.38 &	32.90(+2.14) \\
          10 &Tree-GRU  &tree     &35.67	&34.25	&35.72 &33.47 &	29.95 &	33.35(+2.59)   \\
          11 &Bidirectional     &tree     &\textbf{36.57}    &\textbf{35.64} 	&\textbf{36.63} 	&\textbf{34.35} 	&\textbf{30.57} 	&\textbf{34.30(+3.54)}   \\
\end{tabular}
\caption{BLEU scores of different systems. ``Sequential'', ``Tree-LSTM'', ``Tree-GRU'' and ``Bidirectional'' denote the encoder part for the standard sequential encoder, Tree-LSTM encoder, Tree-GRU encoder and the bidirectional tree encoder, respectively. ``no'', ``word'' and ``tree'' in column ``Coverage'' represents the decoder part for using no coverage (standard attention), word coverage \cite{tu-EtAl:2016:P16-1} and our proposed tree-coverage model, respectively.}
\label{table-results}
\end{table*}

\begin{table*} 
	\centering
	\begin{tabular}{c|l|c|c|cccc|c}
		\multicolumn{1}{c|} {\textbf{\#}} &{\textbf{System}} &{\textbf{Coverage}} &\textbf{MT02}  &\textbf{MT03} &\textbf{MT04}  &\textbf{MT05}  &\textbf{MT06} &\textbf{Average} \\ \hline
          12$'$ &Seq-LSTM    &no       &34.98 &32.81 &34.08 &31.39 &28.03 &31.58(+0.82)    \\
          13$'$ &SeqTree-LSTM &no       &35.28 &33.56 &34.94 &32.64 &29.26 &32.60(+1.84)      \\
\end{tabular}
\caption{BLEU scores of different systems based on LSTM. ``Seq-LSTM'' denotes both the encoder and decoder parts for the sequential model are based on LSTM; ``SeqTree-LSTM'' means using Tree-LSTM encoder on top of ``Seq-LSTM''.}
\label{table-lstm}
\end{table*}

\subsection{Model and Training Details}

We compare our proposed models with several state-of-the-art NMT systems and techniques:
\begin{itemize}
    \item \textbf{NMT}: the standard attentional NMT model~\cite{bahdanau2014neural}.
    \item \textbf{Tree-LSTM}: the attentional NMT model extended with the Tree-LSTM encoder~\cite{eriguchi-hashimoto-tsuruoka:2016:P16-1}.
    \item \textbf{Coverage}: the attentional NMT model extended with word coverage~\cite{tu-EtAl:2016:P16-1}.
\end{itemize}
We used the dl4mt implementation of the attentional model,\footnote{\url{https://github.com/nyu-dl/dl4mt-tutorial}} reimplementing the tree encoder and word coverage models.
The word embedding dimension is 512. The hidden layer sizes of both forward and backward sequential encoder are 1024 (except where indicated). Since our Tree-GRU encoders are built on top of the bidirectional sequential encoder, the size of the hidden layer (in each direction) is 2048. For the coverage model, we set the size of coverage vectors to 50.

We use Adadelta~\cite{zeiler2012} for
optimization using a mini-batch size of 32. All other settings are the same as in \citet{bahdanau2014neural}.

We use case insensitive
4-gram BLEU~\cite{Papineni2002} for  evaluation, as calculated by  \texttt{multi-bleu.perl} in the Moses toolkit.\footnote{\url{http://www.statmt.org/moses}}

\begin{table*}
	\centering
	\begin{tabular}{c|l|c|c|cccc|c}
		\multicolumn{1}{c|} {\textbf{\#}} &{\textbf{Encoder}} &{\textbf{Coverage}} 
		&\textbf{MT02}  &\textbf{MT03} &\textbf{MT04}  &\textbf{MT05}  &\textbf{MT06} &\textbf{Average} \\ \hline
        3$'$&Tree-GRU  &no      
        &34.92 &32.79 &34.16 &32.03 &28.75 &31.93(+1.17)  \\
        4$'$&Bidirectional      &no     
        &35.02   &32.64	&35.04	&32.50	&29.72	&32.48(+1.72)   \\         14$'$&Bidirectional-head &no     
        &34.66   &33.17	&34.78	&31.70	&28.47	&32.03(+1.27)
      \end{tabular}
	\caption{Experiments with 512 hidden units in each direction of the sequential encoder. The bidirectional tree encoder using head-lexicalization (Bidirectional-head), proposed by~\cite{DBLP:journals/corr/TengZ16}, does not work as well as our simpler bidirectional tree encoder (Bidirectional). }
	\label{table-dim}
\end{table*}

\subsection{Tree Encoders}

This set of experiments evaluates the effectiveness of our proposed tree encoders. Table~\ref{table-results}, row 2 confirms the finding of \newcite{eriguchi-hashimoto-tsuruoka:2016:P16-1} that a Tree-LSTM encoder helps, and row 3 shows that our Tree-GRU encoder gets a better result (+0.87 BLEU, v.s. row 2). To verify our assumption that model consistency is important for performance, we also conduct experiments to compare Tree-LSTM and Tree-GRU on top of LSTM-based encoder-decoder settings. Tree-Lstm with LSTM based sequential model can obtain 1.02 BLEU improvement(Table~\ref{table-lstm}, row 13$'$), while Tree-LSTM with GRU based sequential model only gets 0.75 BLEU improvement. Although Tree-Lstm with LSTM based sequential model obtain a slightly better result(+0.22 BLEU, v.s. Table~\ref{table-results}, row 3), it has more parameters(+1.6M) and takes 1.3 times longer for training.

Since the annotation size of our bidirectional tree encoder is twice of the Tree-LSTM encoder, we halved the size of the hidden layers in the sequential encoder to 512 in each direction, to make fair comparison. These results are shown in Table~\ref{table-dim}. Row 4$'$ shows that, even with the same annotation size, our bidirectional tree encoder works better than the original Tree-LSTM encoder (row 2). In fact, our halved-sized unidirectional Tree-GRU encoder (row 3$'$) also works better than the Tree-LSTM encoder (row 2) with half of its annotation size.

We also compared our bidirectional tree encoder with the head-lexicalization based bidirectional tree encoder proposed by~\newcite{DBLP:journals/corr/TengZ16}, which forms the input vector for each non-leaf node by a bottom-up head propagation mechanism (Table~\ref{table-dim}, row 14$'$). Our bidirectional tree encoder gives a better result, suggesting that head word information may not be as helpful for machine translation as it is for syntactic parsing.

When we set the hidden size back to 1024, we found that training the bidirectional tree encoder was more difficult. Therefore, we adopted a two-phase training strategy: first, we train the parameters of the bottom-up encoder based NMT system; then, with the initialization of bottom-up encoder and random initialization of the top-down part and decoder, we train the bidirectional tree encoder based NMT system. Table~\ref{table-results}, row 4 shows the results of this two-phase training: the bidirectional model (row 4) is 0.79 BLEU better than our unidirectional Tree-GRU (row 3).

\subsection{Tree-Coverage Model}

Rows 5--8 in Table~\ref{table-results} show that the word coverage model of \citet{tu-EtAl:2016:P16-1} consistently helps when used with our proposed tree encoders, with the bidirectional tree encoder remaining the best.
However, the improvements of the tree encoder models are smaller than that of the baseline system. This may be caused by the fact that the word coverage model neglects the relationship among the trees, e.g. the relationship between children and parent nodes. Our tree-coverage model consistently improves performance further (rows 9--11).

Our best model combines our bidirectional tree encoder with our tree-coverage model (row 11), yielding a net improvement of +3.54 BLEU over the standard attentional model (row 1), and +1.90 BLEU over the stronger baseline that implements both the bottom-up tree encoder and coverage model from previous work (row 6).

As noted before, the original coverage model does not take word order into account. For comparison, we also implement an extension of the coverage model that lets each coverage vector also depend on those of its left and right neighbors at the previous time step. This model does not help; in fact, it reduces BLEU by about 0.2.

\subsection{Analysis By Sentence Length}

\begin{figure}[t]
	\centering
	\includegraphics[width=.47\textwidth]{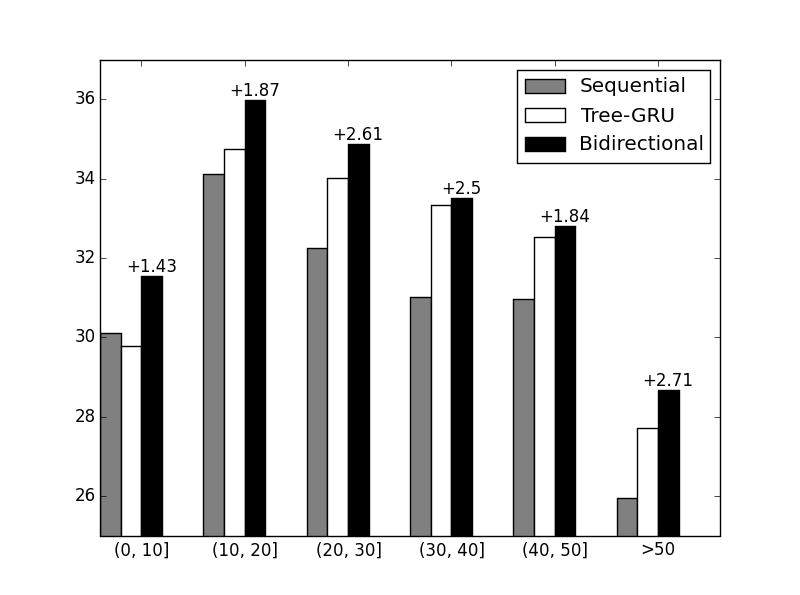}

    \caption{Performance of translations with respect to the lengths of the source sentences. ``+" indicates the improvement over the baseline sequential model.}
	
	\label{fig-top}
\end{figure}

Following~\newcite{bahdanau2014neural}, we bin the development and test sentences by length and show BLEU scores for each bin in Figure~\ref{fig-top}. The proposed bidirectional tree encoder outperforms the sequential NMT system and the Tree-GRU encoder across all lengths. The improvements become larger for sentences longer than 20 words, and the biggest improvement is for sentences longer than 50 words. This provides some evidence for the importance of syntactic information for long sentences.

\section{Related Work}

Recently, many studies have focused on using explicit syntactic tree structure to help learn sentence representations for various sentence classification tasks. For example, \newcite{DBLP:journals/corr/TengZ16} and \newcite{2017arXiv170101811K} extend the bottom-up model to a bidirectional model for classification tasks, using Tree-LSTMs with head lexicalization and Tree-GRUs, respectively. We draw on some of these ideas and apply them to machine translation. We use the representation learnt from tree structures to enhance the original sequential model, and make use of these syntactic information during the generation phase.

In NMT systems, the attention model~\cite{bahdanau2014neural} becomes a crucial part of the decoder model. \newcite{cohn-EtAl:2016:N16-1} and \newcite{feng-EtAl:2016:COLING3} extend the attentional model to include structural biases from word based alignment models. \newcite{yoonkim} incorporate richer structural distributions within deep networks to extend the attention model. Our contribution to the decoder model is to directly exploit structural information in the attention model combined with a coverage mechanism.

\section{Conclusion}

We have investigated the potential of using explicit source-side syntactic trees in NMT by proposing a novel syntax-aware encoder-decoder model.
Our experiments have demonstrated that a top-down encoder is a useful enhancement for the original bottom-up tree encoder~\cite{eriguchi-hashimoto-tsuruoka:2016:P16-1}; and incorporating syntactic structure information into the decoder can better control the translation. Our analysis suggests that the benefit of source-side syntax is especially strong for long sentences.

Our current work only uses the structure part of the syntactic tree, without the labels. For future work, it will be interesting to make use of node labels from the tree, or to use syntactic information on the target side, as well.

\section*{Acknowledgments}
The authors would like to thank the anonymous reviewers for their valuable comments. This work is supported by the
National Science Foundation of China (No. 61672277, 61300158 and 61472183). Part of Huadong Chen's contribution was made when visiting University of Notre Dame. His visit was supported by the joint PhD program of China Scholarship Council.

\bibliography{acl2017}
\bibliographystyle{acl_natbib}

\end{document}